\def\year{2021}\relax
\documentclass[letterpaper]{article} % DO NOT CHANGE THIS
\usepackage{aaai21}  % DO NOT CHANGE THIS
\usepackage{times}  % DO NOT CHANGE THIS
\usepackage{helvet} % DO NOT CHANGE THIS
\usepackage{courier}  % DO NOT CHANGE THIS
\usepackage[hyphens]{url}  % DO NOT CHANGE THIS
\usepackage{graphicx} % DO NOT CHANGE THIS
\usepackage{natbib}  % DO NOT CHANGE THIS AND DO NOT ADD ANY OPTIONS TO IT
\usepackage{caption} % DO NOT CHANGE THIS AND DO NOT ADD ANY OPTIONS TO IT
\usepackage{csquotes}
\MakeOuterQuote{"}
\usepackage{multirow}

\usepackage{microtype}
\usepackage{cite}
\usepackage{amsmath,amssymb,amsthm}
\usepackage{amsfonts}

\usepackage{subcaption} % comment subfigure package while using it
\usepackage{enumitem} % to reduce space between items in enumerate

\def\Vec#1{{\boldsymbol{#1}}} % vectors
\def\Mat#1{{\boldsymbol{#1}}} % martices

\title{Attr2Style: A Transfer Learning Approach for Inferring Fashion Styles via Apparel Attributes}
\author{Rajdeep H Banerjee \textsuperscript{$\dag$,1}, Abhinav Ravi \textsuperscript{$\dag$,2}, Ujjal Kr Dutta \textsuperscript{$\dag$,3} \\
\textsuperscript{$\dag$} Myntra\\
{\small \{\textsuperscript{1}rajdeep.banerjee,\textsuperscript{2}abhinav.ravi,\textsuperscript{3}ujjal.dutta\}@myntra.com}
}

\begin{document}

\maketitle

\begin{abstract}
Popular fashion e-commerce platforms mostly provide details about low-level \textit{attributes} of an apparel (for example, neck type, dress length, collar type, print etc) on their product detail pages. However, customers usually prefer to buy apparel based on their \textit{style information}, or simply put, \textit{occasion} (for example, party wear, sports wear, casual wear etc). Application of a supervised image-captioning model to generate style-based image captions is limited because obtaining ground-truth annotations in the form of style-based captions is difficult. This is because annotating style-based captions requires a certain amount of fashion domain expertise, and also adds to the costs and manual effort. On the contrary, low-level attribute based annotations are much more easily available. To address this issue, we propose a transfer-learning based image captioning model that is trained on a source dataset with sufficient attribute-based ground-truth captions, and used to predict style-based captions on a target dataset. The target dataset has only a limited amount of images with style-based ground-truth captions. The main motivation of our approach comes from the fact that most often there are correlations among the low-level attributes and the higher-level styles for an apparel. We leverage this fact and train our model in an encoder-decoder based framework using attention mechanism. In particular, the encoder of the model is first trained on the source dataset to obtain latent representations capturing the low-level attributes. The trained model is fine-tuned to generate style-based captions for the target dataset. To highlight the effectiveness of our method, we qualitatively and quantitatively demonstrate that the captions generated by our approach are close to the actual style information for the evaluated apparel. A Proof Of Concept (POC) for our model is under pilot at Myntra (www.myntra.com) where it is exposed to some internal users for feedback.
\end{abstract}

\section{Introduction}
\begin{figure}[!htb]
%\vspace{-0.7cm}
\centering
	\begin{subfigure}{0.8\linewidth}
    	\centering
		\includegraphics[trim={0cm 0cm 0cm 0cm},clip,width=\linewidth]{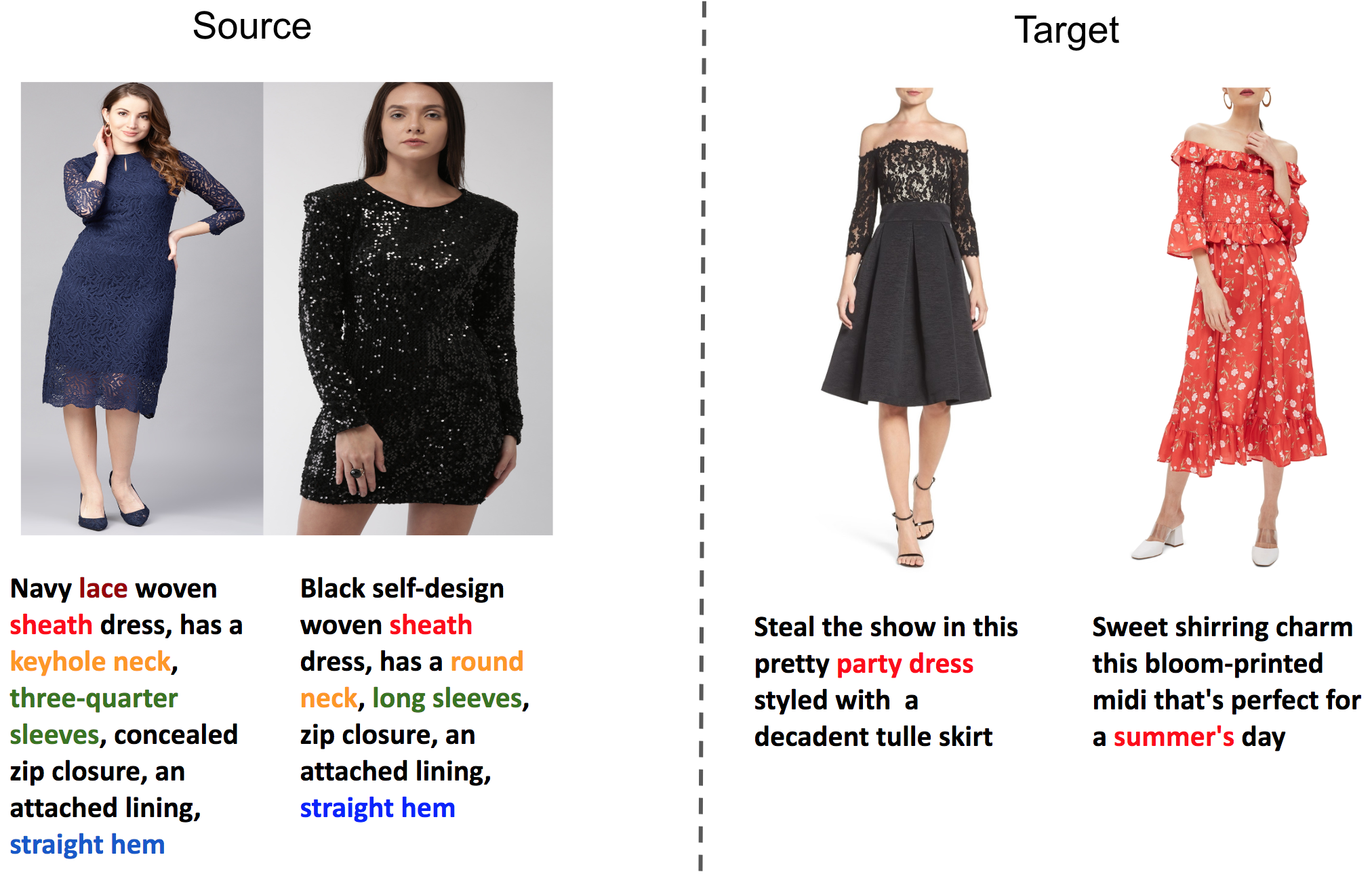}
		\caption{}
        \label{subfig_a}
    \end{subfigure}\\
	\begin{subfigure}{0.8\linewidth}
    	\centering
		\includegraphics[trim={0cm 0cm 0cm 0cm},clip,width=\linewidth]{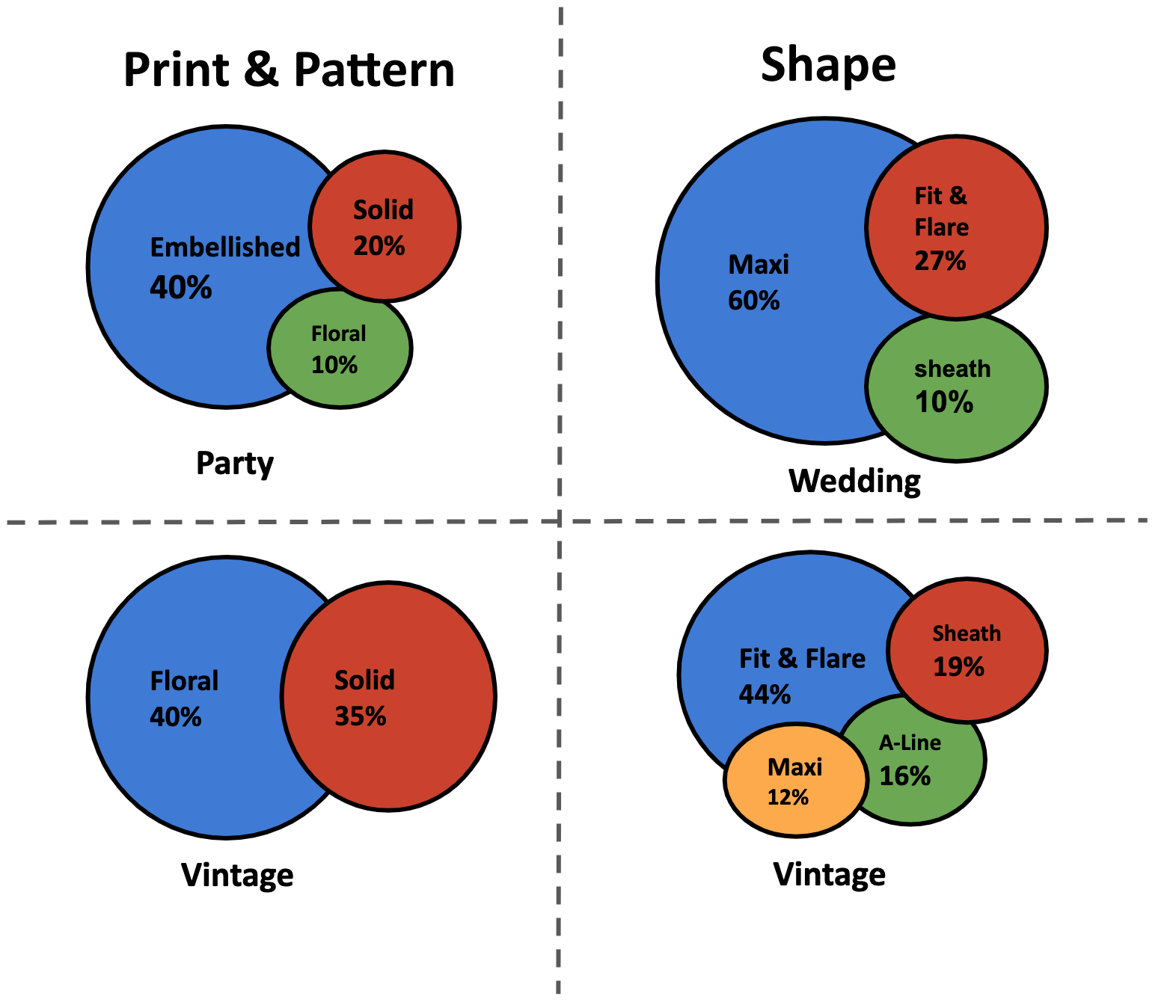}
		\caption{}
        \label{subfig_b}
    \end{subfigure}
    \caption{(a) Illustration of attribute and style based captions from source and target domains respectively, and (b) Correlation among low-level attributes and higher-level style information (for the \textit{vintage} style, left quadrant shows the \textit{print and pattern} based attributes, and the right quadrant shows the \textit{shape} based attributes).}
    %\label{ablation_wt}
%\vspace{-0.9cm}
\end{figure}
\begin{figure*}[!htb]
%\vspace{-1mm}
\centering
	\begin{subfigure}{0.8\linewidth}
    	\centering
		\includegraphics[trim={0cm 0cm 0cm 0cm},clip,width=\linewidth]{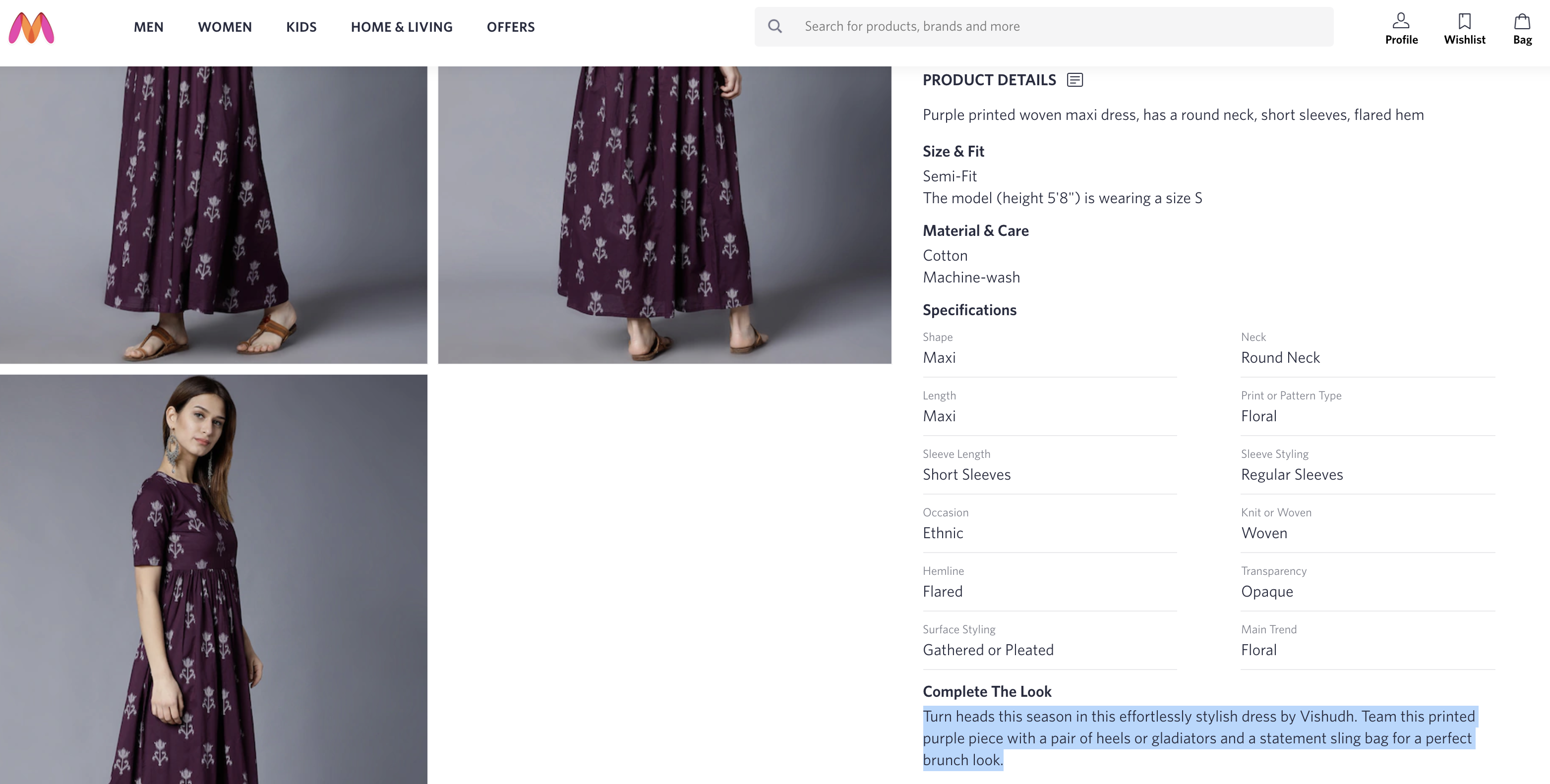}
		\caption{}
        \label{looks_eg1}
    \end{subfigure}\\
	\begin{subfigure}{0.8\linewidth}
    	\centering
		\includegraphics[trim={0cm 0cm 0cm 0cm},clip,width=\linewidth]{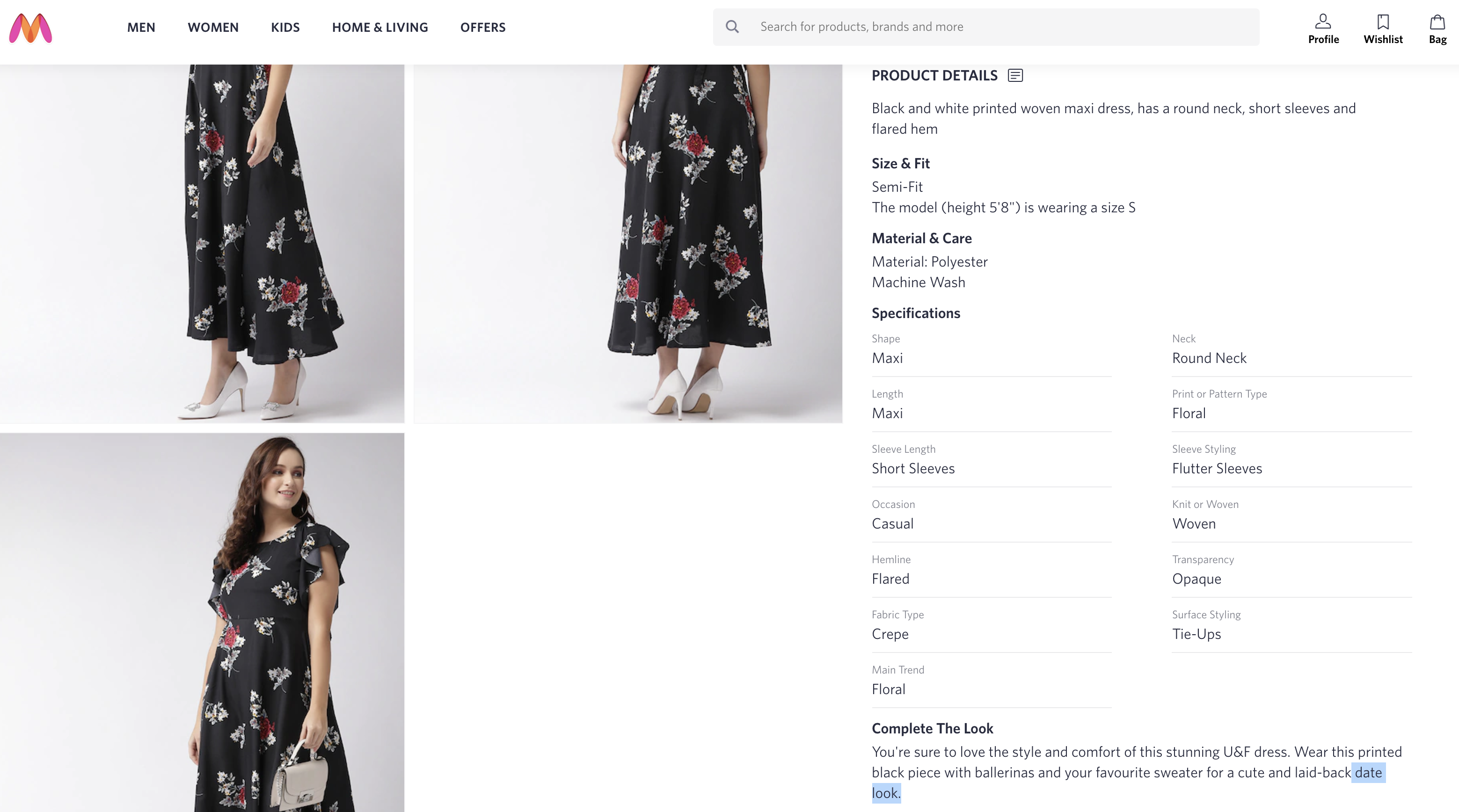}
		\caption{}
        \label{looks_eg2}
    \end{subfigure}
    \caption{Sample product display pages on our platform. a) \textbf{Complete the look} shows a \textit{looks} based caption, with \textit{attribute}-based caption information in \textbf{Product Details}, b) An example of style based caption for the \textit{date look}.}
    \label{looks_egs}
%\vspace{-0.85cm}
\end{figure*}

\begin{figure*}[!htb]
\centering
\includegraphics[width=0.7\linewidth]{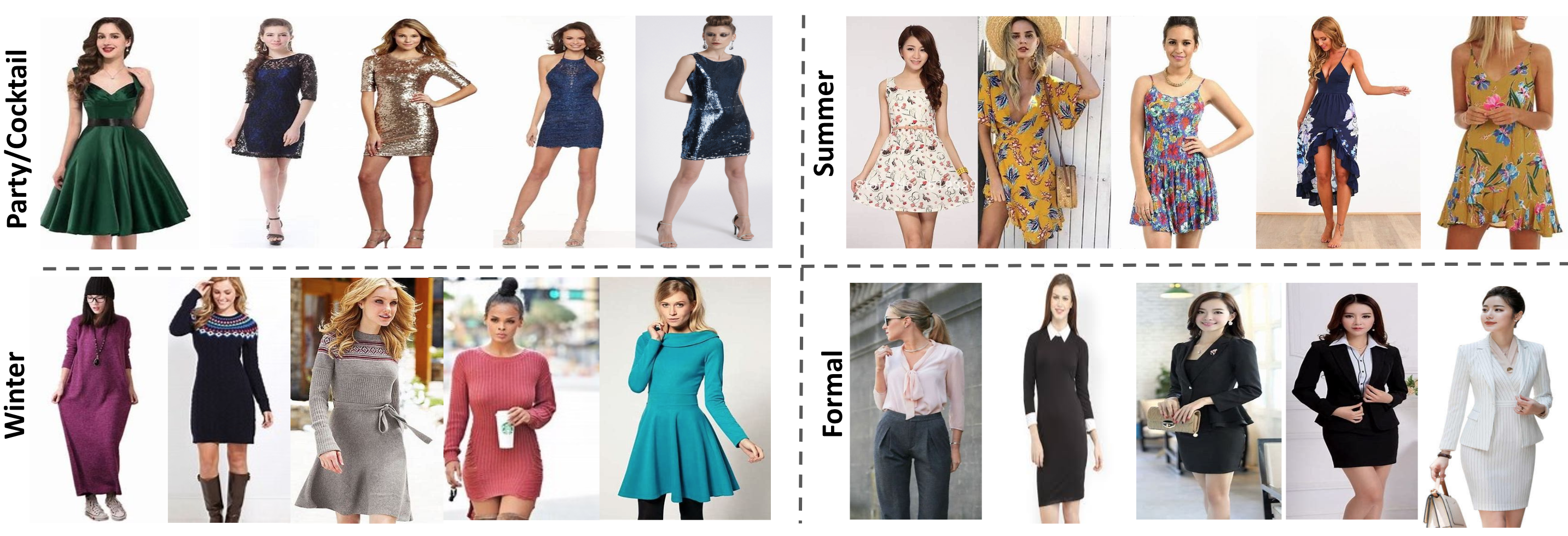}
\caption{Different types of looks.}
\label{looks}
\end{figure*}

\begin{figure*}[!htb]
%\vspace{-1cm}
\centering
\includegraphics[width=0.85\linewidth]{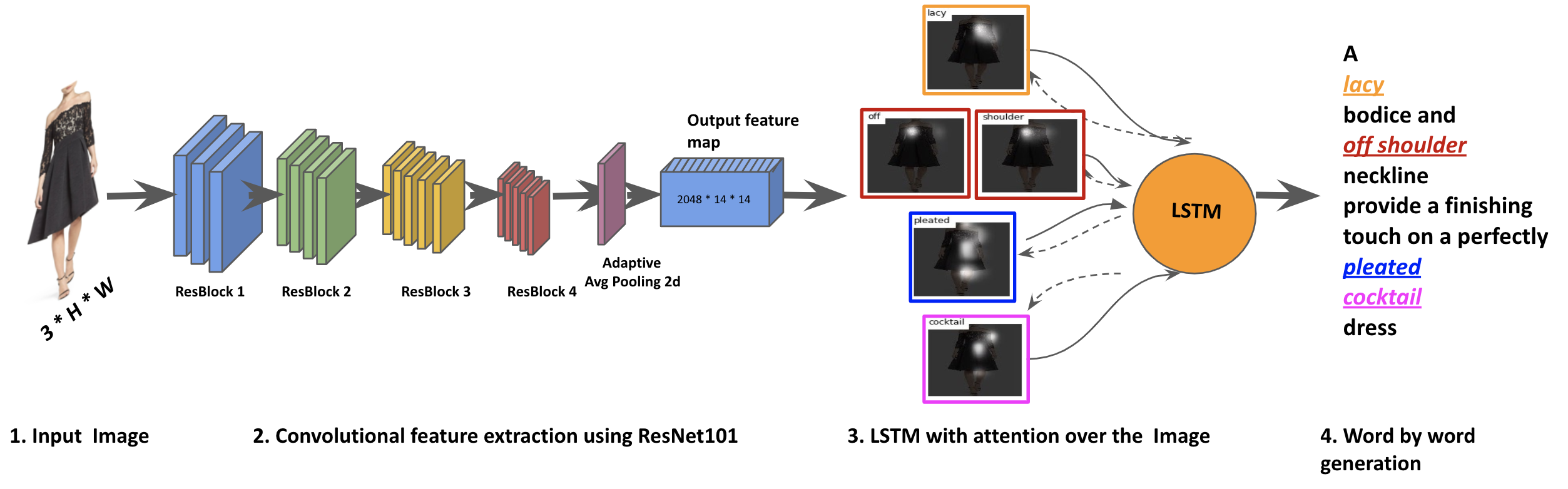}
\caption{Our image caption network\label{network}}
%\vspace{-0.8cm}
\end{figure*}
Catalog images of fashion e-commerce websites are mostly annotated with captions providing details about the low-level attributes of an apparel (for example, neck type, dress length, collar type, print etc). Such captions are easier to annotate as low-level attributes being generic in nature are easier to obtain. Often, apparel manufacturers themselves provide such information. However, captions providing \textit{nature} or \textit{style} (or \textit{looks}) information of an apparel (Figure \ref{subfig_a}) are relatively less common (for example, party wear, sports wear, casual looks etc). This is despite the fact that users have a higher preference for \textit{style} information over \textit{low-level attributes} while buying apparel for occasions. A straightforward solution to address this problem would be to annotate a dataset with \textit{style} information based captions and train an image captioning model. However, annotating style-based captions is not trivial, and requires a certain amount of fashion domain knowledge, in addition to economic expenses and manual efforts.

A clear look at apparel indicates that \textit{attributes} and \textit{styles} often have correlations among them. For example, as shown in Figure \ref{subfig_b}, a high percentage of \textit{party} dresses have \textit{embellished prints} as the dominant attribute, with \textit{floral prints} as the minor one. On the other hand the \textit{vintage} style has the \textit{floral prints} as the dominant attribute. Due to this observation, we conjecture that the lack of style-based ground truth captions for images could be addressed by a transfer learning approach via attribute-based information.

In this paper, we propose an innovative AI system that addresses the above issue by transfer learning. In particular, we apply our method to overcome the scarcity of style-based image captions on our own fashion e-commerce platform called Myntra (www.myntra.com). Figure \ref{looks_egs} shows sample images of Product Display Pages (PDP) on our platform, showing both types of captions, and Figure \ref{looks} shows common types of \textit{looks}. However, we would like to note that the style-based captions are available for only a handful of images, whereas there is an abundance of images with attribute-based captions. To learn the style-based captions via images with attribute-based captions, we cast our problem into a transfer learning setting. Specifically, we consider the large set of images with attribute-based captions as a \textit{source domain dataset}, and the handful of images with style-based captions as a \textit{target domain dataset}.

Firstly, we train a machine learning model on the \textit{source domain dataset} to learn latent information corresponding to the attributes, and then transfer this knowledge to that of the \textit{target domain dataset} to learn information corresponding to the styles (i.e., looks). To this end, we propose an attention-based image captioning model with an encoder-decoder that leverages transfer learning to obtain style-based captions for target domain images. We first train our model on a source dataset which has abundant attribute-based ground truth captions. The encoder of the trained model now captures the attribute information in the form of latent embeddings. The model is then fine-tuned on the target dataset, which has only a limited number of images with ground truth style-based captions. By virtue of the latent representations learned by the encoder, we are able to \textit{transfer} knowledge of the attributes from the source domain and learn better style-based captions for images from the target domain.

\textbf{Discussion on the transfer method:} The transfer mechanism of our proposed approach is similar in spirit to that of the Domain-Adversarial training of Neural Networks (DANN) method \cite{ganin2016domain}, that studied a representation learning approach for domain adaptation. Their approach is directly inspired by the theory on domain adaptation suggesting that, for effective domain transfer to be achieved, predictions must be made based on \textit{features} (that cannot discriminate between the source and target domains). They make use of an adversarial loss to learn \textit{latent features}. Their features are very generic in nature. However, in our case, as shown in Figure \ref{subfig_b}, there is a well-known correlation between lower level attributes and higher level styles. Thus, the encoder of the network trained on attribute level captions produces \textit{latent representations} which are agnostic to the domain knowledge (thus, satisfying the theory on domain adaptation). Now this same latent representation helps in \textit{transferring} the low-level attribute information to the higher level styles, and hence act as good representations for caption generation. Despite the \textit{two-phase like training}, in principle, transfer learning is achieved by virtue of the latent features learned. Our approach is indeed a crafty and subtle way of performing transfer learning. 

\section{Proposed Method}

\textbf{Model architecture:} Figure \ref{network} illustrates our proposed encoder-decoder based image captioning model, that consists of the following major components: i) An encoder, wherein we make use of a ResNet101 \cite{he2016deep} (pretrained on Imagenet \cite{deng2009imagenet}) to obtain the \textit{latent representations} (that help in \textit{transfer learning}), and ii) A decoder (a LSTM network \cite{hochreiter1997lstm}), that makes use of the latent features to provide image captions. We incorporate an attention mechanism in the decoder to obtain a correspondence between the feature vectors and portions of the 2-D image. For this, we extract features from a lower convolution layer of the network, hence allowing the decoder to selectively focus on certain parts of an image (soft attention as in \cite{xu2015show}). The ResNet based encoder has the option of fine-tuning convolution blocks 2 through 4. The final encoding produced by our ResNet101 encoder has a size of $14\times 14$ with $2048$ channels.
\begin{figure*}[!htb]
%\vspace{-1mm}
\centering
	\begin{subfigure}{0.6\linewidth}
    	\centering
		\includegraphics[trim={0cm 0cm 0cm 0cm},clip,width=\linewidth]{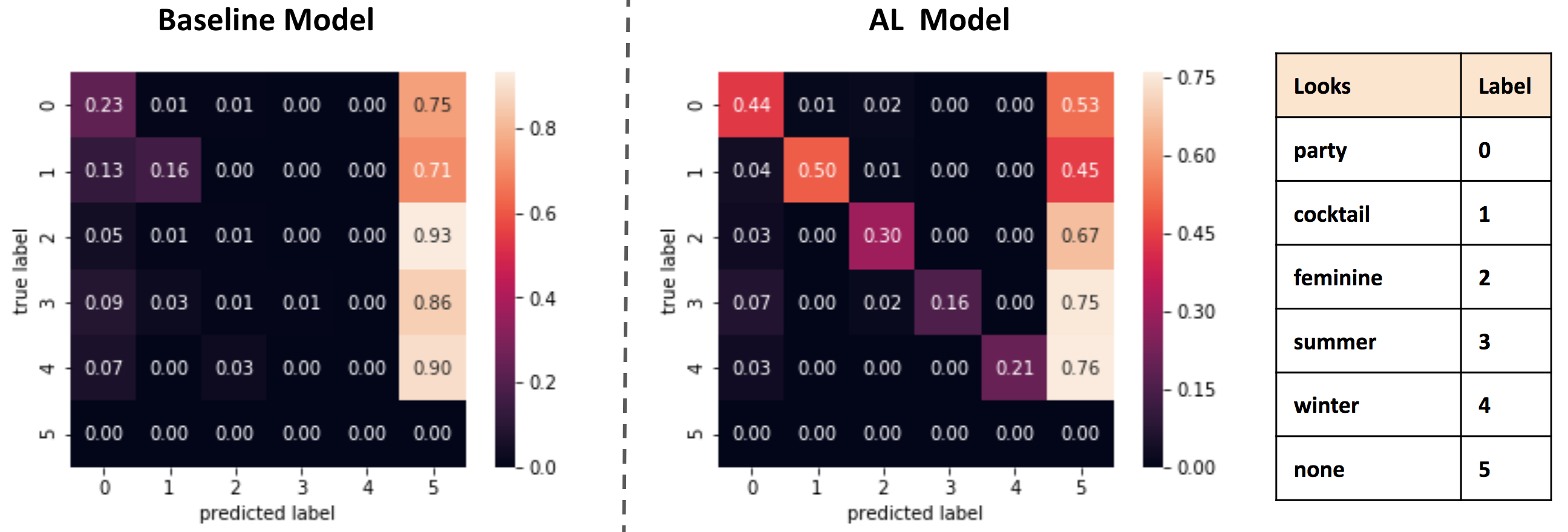}
		\caption{}
        \label{conf_mat}
    \end{subfigure}\\
	\begin{subfigure}{0.6\linewidth}
    	\centering
		\includegraphics[trim={0cm 0cm 0cm 0cm},clip,width=\linewidth]{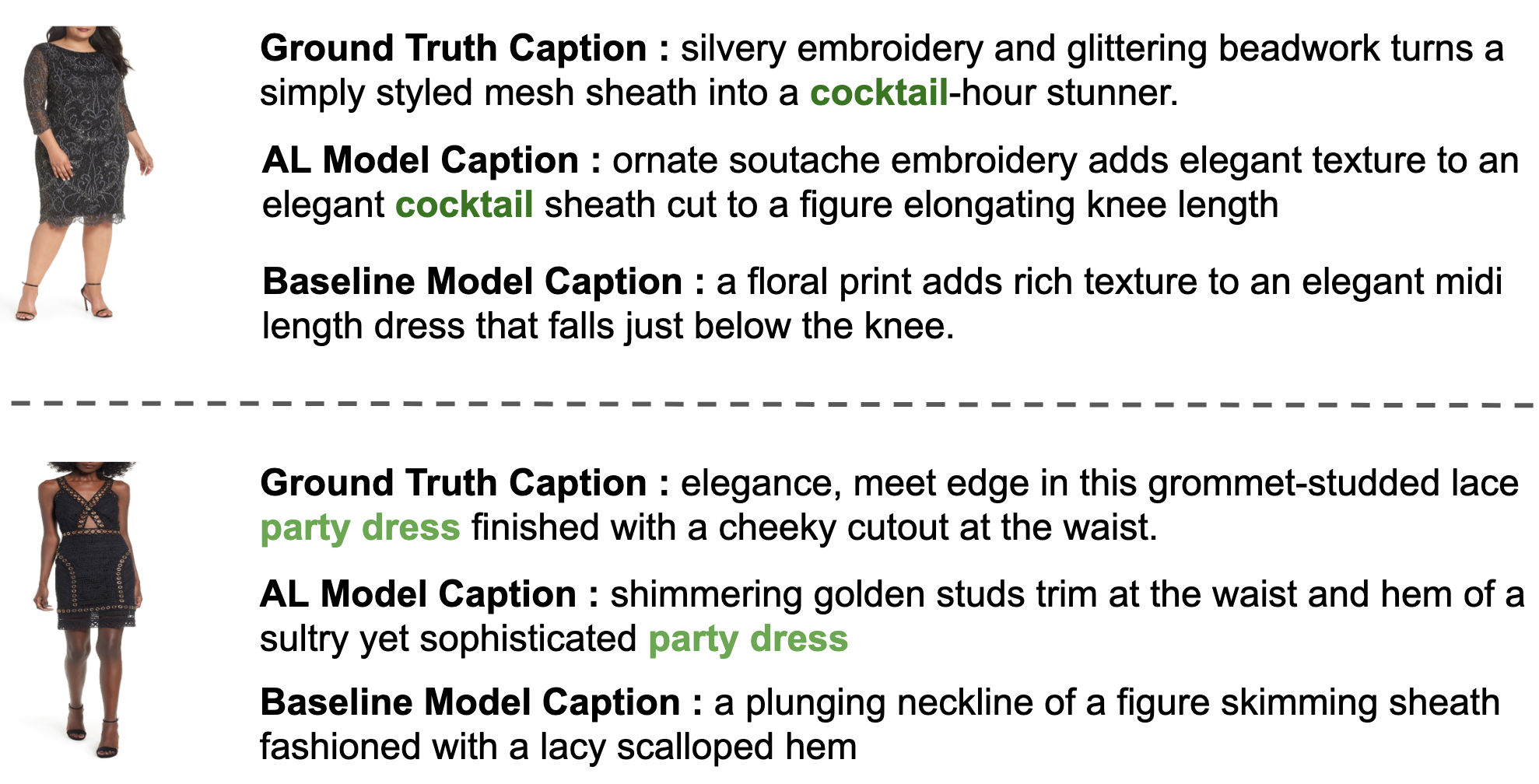}
		\caption{}
        \label{captions_target_ds}
    \end{subfigure}
    \caption{(a) Confusion matrices for the baseline and our method, and (b) Comparison of generated captions using the baseline method and our method.}
    %\label{ablation_wt}
%\vspace{-0.4cm}
\end{figure*}

\begin{figure*}[!htb]
%\vspace{-1mm}
\centering
	\begin{subfigure}{0.6\linewidth}
    	\centering
		\includegraphics[trim={0cm 0cm 0cm 0cm},clip,width=\linewidth]{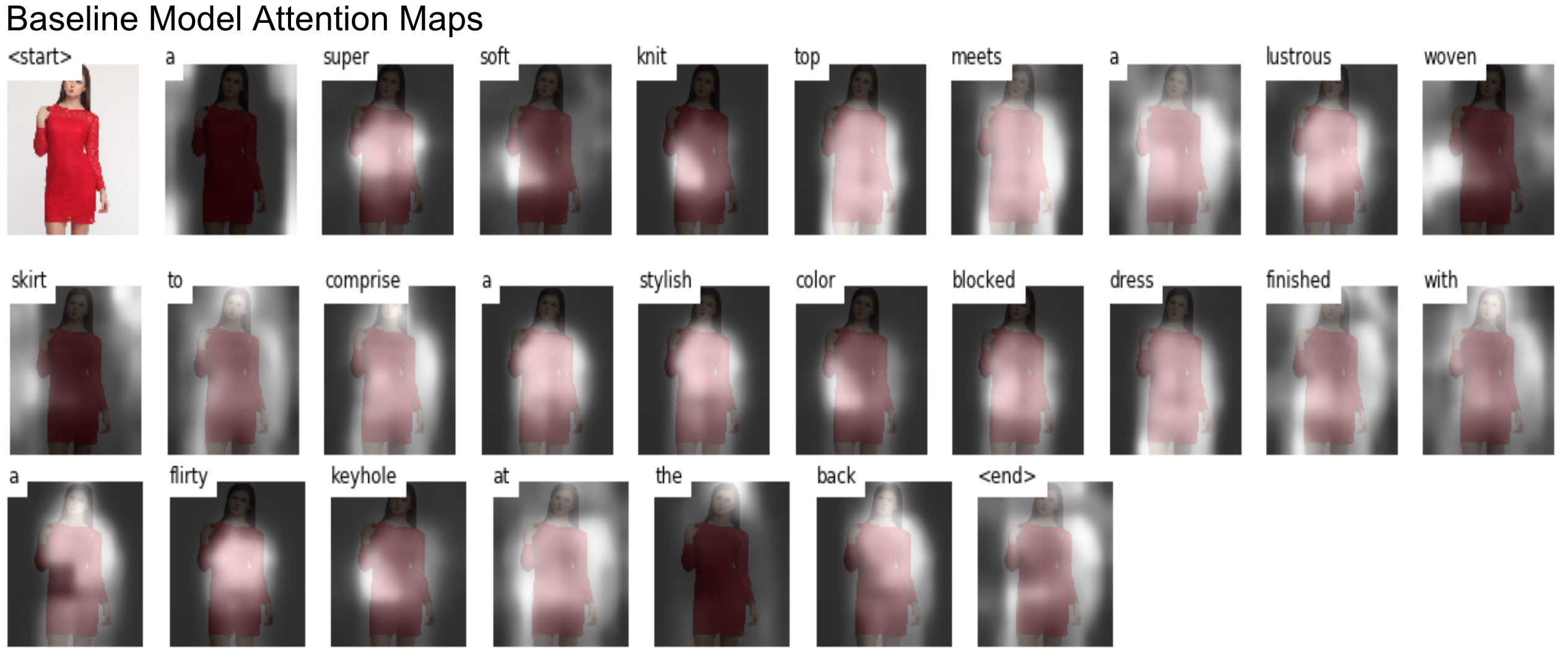}
		\caption{}
        \label{attention_maps_baseline}
    \end{subfigure}\\
	\begin{subfigure}{0.6\linewidth}
    	\centering
		\includegraphics[trim={0cm 0cm 0cm 0cm},clip,width=\linewidth]{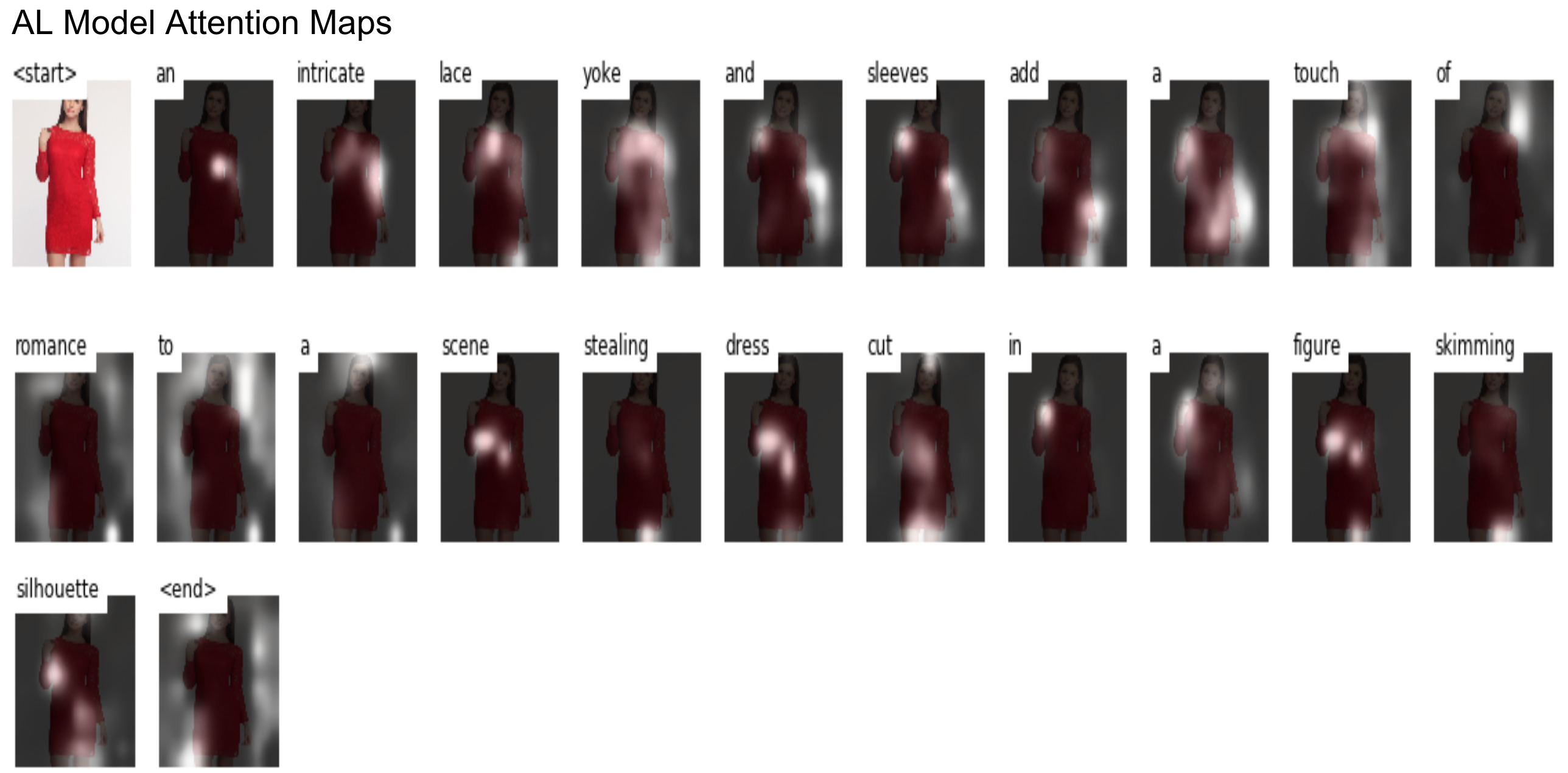}
		\caption{}
        \label{attention_maps_ours}
    \end{subfigure}
    \caption{(a) Attention maps of captions using the baseline, and (b) Attention maps of captions by our AL-model.}
    \label{attention_maps}
%\vspace{-0.85cm}
\end{figure*}

\textbf{Attention-based LSTM model}
For a pair of sequences $\Mat{A}=\{w_1^a,\cdots,w_m^a\}$ and $\Mat{B}=\{w_1^b,\cdots,w_n^b\}$, let their LSTM encoding be denoted as:
\begin{equation}
\begin{split}
    &\Vec{h}_t^a=\textrm{lstm}(e(w_t^a),\Vec{h}_{t-1}^a), \forall t \in [1,m]\\
    &\Vec{h}_t^b=\textrm{lstm}(e(w_t^b),\Vec{h}_{t-1}^b), \forall t \in [1,n]
\end{split}
\end{equation}
Here, $e(w)$ is the corresponding embedding for the word $w$. The last hidden state $\Vec{h}_n^b \in \mathbb{R}^d$ is used to \textit{attend} the intermediate representations $\Mat{H}^a=\{\Vec{h}_1^a ,\cdots, \Vec{h}_m^a\} \in \mathbb{R}^{m \times d}$ of $\Mat{A}$, using the attention-mechanism \cite{bahdanau2014neural}:
\begin{equation}
\begin{split}
    &\Tilde{\alpha}_t = \Vec{v}^\top \textrm{tanh} (\Mat{W}_1\Vec{h}_t^a+\Mat{W}_2\Vec{h}_n^b+\Vec{b}), \forall t \in[1,m] \\
    & \alpha_t = \textrm{softmax}(\Tilde{\alpha}_t) \\
    & \Vec{c}_{\alpha}=\sum_{t=1}^m \alpha_t \Vec{h}_t^a
\end{split}
\end{equation}
Here, $\Mat{W}_1 \in \mathbb{R}^{d1 \times d}$, $\Mat{W}_2 \in \mathbb{R}^{d1 \times d}$, $\Vec{b}\in \mathbb{R}^{d1}$ and $\Vec{v}\in \mathbb{R}^{d1}$ are parameters to be learned, and $\Vec{c}_{\alpha}$ is called the \textit{context vector}.

In our case, the LSTM network generates captions one word at every time step $t$ conditioned on the context vector $\Vec{c}_{\alpha}$, the previous hidden state $\Vec{h}_{t-1}$ and the previously generated words. The context vector $\Vec{c}_{\alpha}$ is a dynamic representation of the relevant part of the image input at time $t$.

Let, $\mathbf a_i$, $i=1,...,L$ denote annotation vectors that are features extracted at different image locations. Using the soft-attention mechanism discussed as above, we can redefine our context vector $\Vec{z}_t$ as:
\begin{equation}
\Vec{z}_t = \sum_{i} \alpha_{t,i} \mathbf a_i
\end{equation}
Here, we have,
\begin{equation}
    \alpha_{t,i} = \frac{exp(e_{ti})}{\sum_{k=1}^{L} exp(e_{tk})}
\end{equation}
\begin{equation}
    e_{ti} = f_{att}(\mathbf a_i,\mathbf h_{t-1})
\end{equation}
where f$_{att}$ is a multilayer perceptron.

\textbf{Training:} We first train the caption generator network using attention to generate \textit{attribute captions} using the source dataset. We use the learned image encoder weights to fine tune the network for generating \textit{style-based captions} using the target dataset.

\section{Experiments}
% Here we describe the dataset
To evaluate the proposed method, we made use of images from our fashion e-commerce website \textit{{www.myntra.com}}. As the source dataset, we collected a subset of 20000 images that have captions providing \textit{low-level attribute} information (but no \textit{style-based captions}). We collected another limited small subset of 2500 images for which we already had in-house annotated captions describing the \textit{style information}. This second subset is considered as the target dataset, for which we do not make use of the attribute based annotations. We make use of a distinct set of $430$ test images (with ground truth style-captions) for evaluating the generated captions. The test data has the following \textit{styles}: party, cocktail, feminine, summer, winter, and none (for rest of images).

\textbf{Models compared:} To demonstrate the effectiveness of the proposed \textit{transfer learning} based approach, we conduct an experiment comparing two models: i) \textbf{Baseline Model:} We directly use the available labeled data from the target dataset (with style-based annotations) and train an image captioning model end-to-end using the same architecture as ours (with ImageNet based pretrained weights), for 30 epochs. ii) \textbf{AL-model:} The Attribute-Looks model(AL-model) refers to our proposed method. Essentially, we first train our model using the labeled data from the source dataset (with attribute based annotations) in an end-to-end fashion for 30 epochs. Now, we use the same weights for the trained encoder, and fine-tune the model using the limited labeled data from the target dataset (with style-based annotations) for 30 epochs.
% \begin{figure}[!htb]
% \centering
% \includegraphics[width=0.5\columnwidth]{figures/final_cm.png}
% \caption{Confusion Matrix for generated looks (Baseline vs AL Model)\label{cm}}
% \end{figure}

% \begin{figure}[!htb]
% \centering
% \includegraphics[width=0.6\columnwidth]{figures/mynt_caption.png}
% \caption{Source dataset generated captions\label{mynt_captions}}
% \end{figure}

% \begin{figure}[!htb]
% \centering
% \includegraphics[width=0.6\columnwidth]{figures/nord_captions.png}
% \caption{Target dataset generated captions\label{nord_captions}}
% \end{figure}

% \begin{figure*}[!htb]
% \centering
% \includegraphics[width=0.6\columnwidth]{figures/bsline_attn_maps.png}
% \caption{Attention Map of captions using baseline Model\label{attention_maps_bsl}}
% \end{figure*}

% \begin{figure*}[!htb]
% \centering
% \includegraphics[width=0.6\columnwidth]{figures/al_attn_maps.png}
% \caption{Attention Map of captions using AL- Model\label{attention_maps_al}}
% \end{figure*}

% \usepackage{graphicx}
\begin{table}[!htb]
\centering
\caption{Quantitative comparison of our method against the baseline in terms of precision and recall for various looks.}
\label{quant_vs_baseline}
\resizebox{0.7\linewidth}{!}{%
\begin{tabular}{|c|cc|cc|}
\hline
         & \multicolumn{2}{c|}{Precision}  & \multicolumn{2}{c|}{Recall} \\ \hline
Look     & \textbf{Ours}  & Baseline       & \textbf{Ours}   & Baseline  \\ \hline
Party    & \textbf{72.1}  & 40.3           & \textbf{44.0}   & 23.0      \\
Cocktail & \textbf{98.0}  & 76.1           & \textbf{50.0}   & 16.0      \\
Feminine & \textbf{85.7}  & 16.6           & \textbf{30.0}   & 1.0       \\
Summer   & \textbf{100.0} & \textbf{100.0} & \textbf{16.0}   & 1.0       \\ \hline
\end{tabular}%
}
\end{table}

% \usepackage{multirow}
% \usepackage{graphicx}
\begin{table}[!htb]
\centering
\caption{Quantitative comparison of our method against the baseline in terms of BLEU score and overall accuracy. }
\label{quant_bleu_acc}
\resizebox{0.7\linewidth}{!}{%
\begin{tabular}{|c|cc|cc|}
\hline
\multirow{2}{*}{} & \multicolumn{2}{c|}{\textbf{BLEU}} & \multicolumn{2}{c|}{\textbf{Accuracy}} \\ \cline{2-5} 
                  & \textbf{Ours}      & Baseline      & \textbf{Ours}        & Baseline        \\ \hline
Overall           & \textbf{0.29}      & 0.26          & \textbf{0.32}        & 0.08            \\ \hline
\end{tabular}%
}
\end{table}

\textbf{Results:} Figure \ref{conf_mat} shows the empirical performance of both the approaches, using confusion matrices, and the corresponding per \textit{look} precision and recall values are reported in Table \ref{quant_vs_baseline}. In Table \ref{quant_bleu_acc}, we also report the BLEU score to quantify the quality of image captions, and the accuracy over all the looks. A higher value of both these metrics indicates a better captioning performance, and both of these are computed in the range of $0-1$.

For a particular test image, we have a ground truth caption corresponding to one of the ground truth \textit{styles} (eg, party, cocktail etc). We make use of our model to predict caption for a test image, and obtain the predicted \textit{style}. Please note that the predicted style is inferred from the generated caption based on the presence of style key words in the caption (eg, party, cocktail etc). Using the ground truth and predicted styles for the set of test images, we can compute the performance metrics like precision, recall, and accuracy using standard definitions in a multi-class classification setting. For eg, to calculate accuracy, we can use the following formula: (TP+TN)/(TP+TN+FP+FN). Here, TP: True Positive, TN: True Negative, FP: False Positive, and FN: False Negative. Note that in Table \ref{quant_vs_baseline} we report the \textit{class-wise} Precision and Recall values (in this case a class refers to a \textit{style}). In Table \ref{quant_bleu_acc}, we report the accuracy across all the classes.

The best performing method is shown in \textbf{bold}. The superiority of our proposed AL-model highlights the benefit of transfer learning employed by our model. Figure \ref{captions_target_ds} compares the captions generated by the baseline and our proposed method. The generated captions by our method are closer to the ground truth. We also show the attention maps corresponding to both the baseline and our approach, in Figure \ref{attention_maps}. We observed that the maps are more focused to specific regions of an image in our method. This eventually leads to the generation of better quality image captions by our method.

\textbf{Further details of the system}: Following are some of the brief details of the proposed system: i) The hardware used was a Nvidia Tesla V-100 GPU, where the training time was around 2 days. ii) Our model was built for the \textit{dresses} article type on our platform. iii) A pilot was run for around 8-9 months, based on which we figured out that this is a category of focus / emerging category in our geography. Also, studies have revealed that women consumers engage more when styling details are present in the Product Display Page (PDP).

\section{Related Work}
Before concluding the paper, we would also like to highlight some important developments in the probem of image captioning. It is a challenging problem that spans concepts from across image understanding as well as natural language generation. Encoder-decoder based deep models (as used in our paper) have shown state-of-the-art performance in image captioning [Yao et al., 2018]. In general, the de-facto approach is to exploit a Convolutional Neural Network (CNN) (eg., ResNet \cite{ResNet}) to encode image features, followed by a Recurrent Neural Network (RNN) (eg., LSTM \cite{hochreiter1997lstm}) to generate the caption statements with attention mechanism \cite{xu2015show}.

Considerable efforts have been made to obtain specific improvements. \cite{anderson2018bottom} incorporated object-oriented representations. By making use of textual attributes and image regions, some recent works have addressed the problem from a cross-modal point of view \cite{liu2019aligning,liu2018simnet}. However, these approaches instead of comprehending general correlations among image parts, consider an image as unrelated parts \cite{brendel2018approximating,geirhos2018imagenet}.

Some recent works learn a visual relationship and context-aware attention for image captioning \cite{wang2020learning}. Recently \cite{pan2020x}, presented a novel model to capture the second order interactions with channel-wise and spatial bilinear attention. \cite{cornia2020meshed} proposed a novel Transformer-based image captioning architecture.

However, as observed, the discussed approaches aim at solving specific problems pertaining to the general image captioning problem by following sophisticated schemes. On the other hand, in our method we wanted to emphasize the transfer learning component, rather than addressing the advancement of image captioning. For this reason, we do not use a sophisticated image captioning model. To the best of our knowledge, our application of the image captioning model to perform transfer learning in the fashion application discussed in our paper is novel on its own. \textbf{The major contribution of the paper can be seen as the study of the \textit{correlation among the low-level attributes and higher level styles}, and its empirical establishment, by means of both qualitative as well as quantitative observations.} We believe, this could open up newer studies to further leverage such correlations present within fashion article information.

\section{Conclusions}
In this paper, we propose a simple, yet effective, transfer learning based approach to address the issue of style-based image captioning for a target dataset. We employ an attention-based image captioning model using an encoder-decoder to obtain style-based captions for an apparel. Because of the correlation among low-level attributes and higher-level style of an apparel, we first train the model on a source dataset with attribute-based ground truth captions. The latent representations obtained by the encoder helps in transfer learning of attribute information to the higher level style-based caption generation. We establish this fact by comparing our model with another version of our model with the same architecture, but without pretraining on the source dataset with attribute information. The captions generated by our model are closer to the actual ground truths, thus showing the benefit of transfer learning. Our method could be used to provide additional style-based captions for fashion apparel, thus improving the overall customer experience, and possibly increasing add-to-cart ratio.

\section{Acknowledgement}
We would like to thank our colleague Anoop KR for his contribution in reviewing the core of our work and sharing his insights. We are also grateful to our manager Ravindra Babu for his support and in being a source of inspiration and encouragement throughout the project.

%\clearpage
\bibliography{attr2style_IAAI21}

\end{document}